\documentclass[journal,transmag]{IEEEtran}
\usepackage{amsmath,amssymb} 
\usepackage[section]{algorithm} 
\usepackage{algorithmic}
\usepackage{epsfig}
\usepackage{acronym}
\usepackage{bm}
\usepackage{epstopdf}
\usepackage{multirow}
\usepackage{hyperref}
\usepackage{url}
\usepackage{color}
\ifCLASSINFOpdf
\else
\fi
\begin{document}
%
\title{Person Re-identification in Aerial Imagery
}

\author{\IEEEauthorblockN{Shizhou Zhang\IEEEauthorrefmark{1*},
Qi Zhang\IEEEauthorrefmark{1*},
Yifei Yang\IEEEauthorrefmark{1},
Xing Wei\IEEEauthorrefmark{2$\dagger$},
Peng Wang\IEEEauthorrefmark{1},
\\
Bingliang Jiao\IEEEauthorrefmark{1}
and
Yanning Zhang\IEEEauthorrefmark{1},~\IEEEmembership{Senior Member,~IEEE}}
\IEEEauthorblockA{\IEEEauthorrefmark{1}National Engineering Laboratory for Integrated Aero-Space-Ground-Ocean Big Data Application Technology, \\
School of Computer Science and Engineering, Northwestern Polytechnical University, Xi'an 710072, China.}
\IEEEauthorblockA{\IEEEauthorrefmark{2}Faculty of Electronic and Information Engineering, Xi'an Jiaotong University, Xi'an 710049, China.}
\thanks{* indicates co-first author.
$^\dagger$Corresponding author: Xing Wei}}

\markboth{}%
{Shell \MakeLowercase{\textit{et al.}}: Bare Demo of IEEEtran.cls for IEEE Transactions on Magnetics Journals}
%



\IEEEtitleabstractindextext{%
\begin{abstract}
Nowadays, with the rapid development of consumer Unmanned Aerial Vehicles (UAVs), visual surveillance by utilizing the UAV platform has been very attractive. Most of the research works for UAV captured visual data are mainly focused on the tasks of object detection and tracking. However, limited attention has been paid to the task of person Re-identification (ReID) which has been widely studied in ordinary surveillance cameras with fixed emplacements.
In this paper, to facilitate the research of person ReID in aerial imagery, we collect a large scale airborne person ReID dataset named as Person ReID for Aerial Imagery (PRAI-1581), which consists of 39,461 images of 1581 person
identities. The images of the dataset are shot by two DJI consumer UAVs flying at an altitude ranging from 20 to 60 meters above the ground, which covers most of the real UAV surveillance scenarios.
In addition, we propose to utilize subspace pooling of convolution feature maps to represent the input person images. Our method can learn a discriminative and compact feature representation for ReID in aerial imagery and can be trained in an end-to-end fashion efficiently. We conduct extensive experiments on the proposed dataset and the experimental results demonstrate that re-identify persons in aerial imagery is a challenging problem, where our method performs favorably against state of the arts. Our dataset can be accessed via \url{https://github.com/stormyoung/PRAI-1581}.
\end{abstract}

}

\maketitle

\IEEEdisplaynontitleabstractindextext

%
\IEEEpeerreviewmaketitle

\section{Introduction}
%
%
%
%
\IEEEPARstart{P}{erson} Re-identification (ReID) aims to re-identify the same individual from other non-overlapping cameras~\cite{li2014deepreid,wang2015zero,ye2016person,xiao2016learning,zheng2016person,hermans2017defense,zhou2017large,zhang2018pedestrian,chen2018person,wang2019incremental,ding2018feature,zhang2019person,diangang2020}, which is a key technique to achieve the tasks of long-term person tracking, cross camera tracking, person matching \textit{etc}. in the field of intelligent visual surveillance.
Nowadays, with the rapid development of the consumer Unmanned Aerial Vehicles (UAVs),
visual surveillance by utilizing the UAV platform has been booming and become a necessary supplementary tactic for the tradition surveillance situation with fixed camera emplacements.

Recently, more and more efforts have been devoted into the study of intelligent aerial surveillance~\cite{xia2018dota,zhu2018vision,mueller2016benchmark,cheng2016learning} from both the industrial and academia.
However, most of the research works are mainly focused on the tasks of object detection~\cite{zhou2018scale,zhang2018w2f,zhou2018learning} and tracking~\cite{xiang2014monocular}. There is limited attention paid to the task of person ReID,
which is probably because there does not exist a large scale publicly available person ReID dataset facing to the real UAV surveillance scenarios,
although ReID is as important as detection and tracking in UAV-based visual surveillance. As well as we all know, the annotation of the ReID task is much labor-intensive as one needs to annotate not only the bounding box positions, but also the ID numbers of each person.

\begin{figure*}[htb!]
  \centering
  \includegraphics[width=0.9\linewidth]{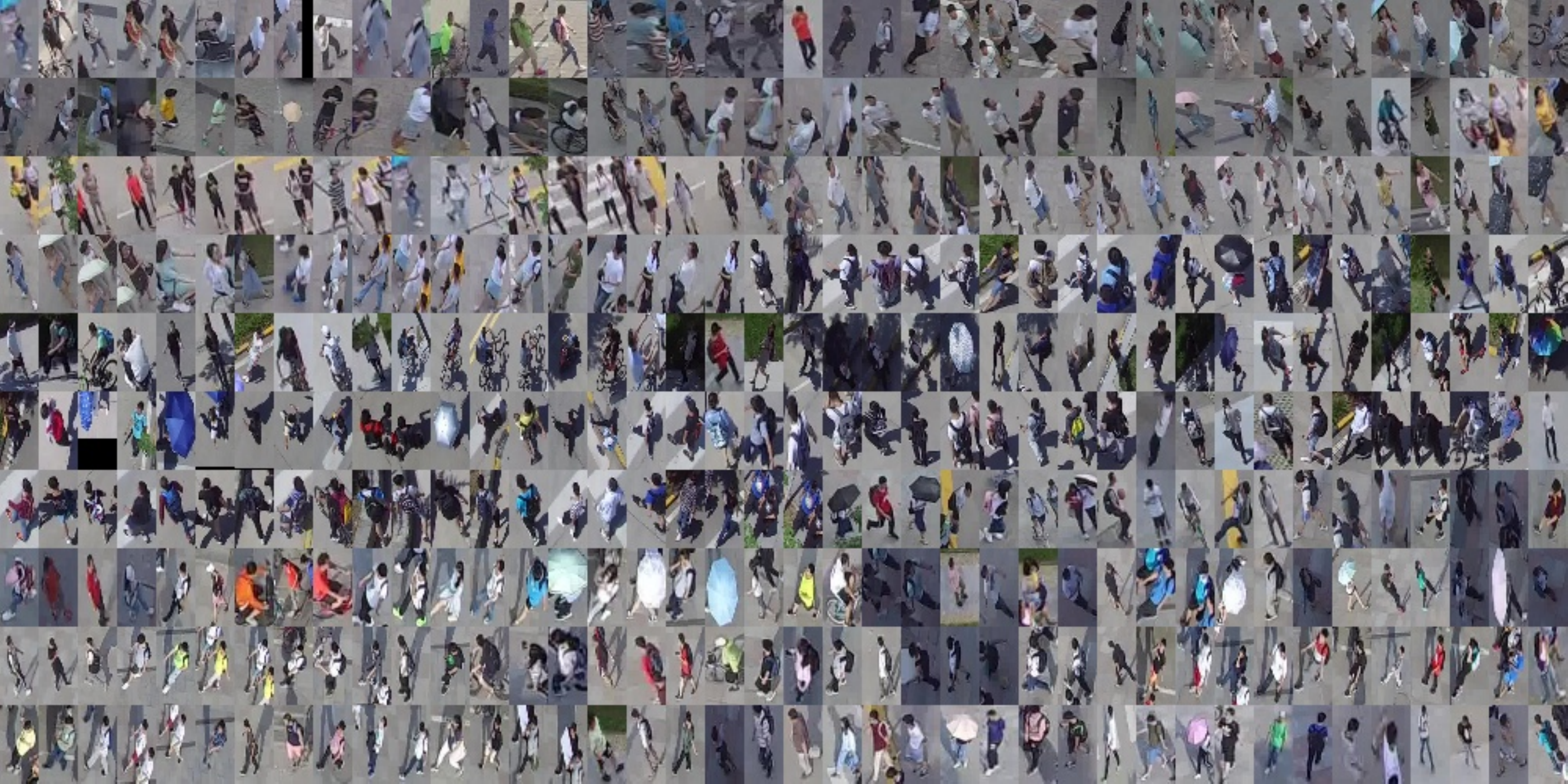}\\
  \caption{Examplar images in our proposed PRAI-1581 dataset.}\label{fig_example_images}
\end{figure*}

In this paper, we collect a large scale airborne person ReID dataset named as Person ReID in Aerial Imagery (PRAI-1581), which consists of 39,461 images of 1581 person identities. The images of the dataset are shot by two DJI consumer UAVs flying at
an altitude ranging from 20 to 60 meters above the ground, which covers most of the real UAV surveillance scenarios. The two UAVs are controlled by two different pilots to fly in different areas with non-overlapping monitoring field. It is worth noting that, due to the variable altitude of the flight, the adjustable tilt angle of the camera, and the free rotation of the fuselage, persons have diverse resolutions, viewpoints and poses in a single UAV, and the entirety scenarios are even more complicated since we have two UAVs controlled independently.
Compared to the person ReID task in ordinary surveillance scenario with fixed camera emplacement, it is much more challenging to match persons in aerial imagery. Figure~\ref{fig_example_images} shows some example images of our collected dataset.
The collection and annotation of our dataset takes us more than 800 hours with 10 well-trained annotators.
Besides the person ReID task, our dataset can also be used as a benchmark dataset for person detection and tracking in aerial images.
We will release our dataset and benchmark to the public to facilitate researches in this area.

On the other hand, learning discriminative and invariant features is the key to visual recognition, especially for the ReID task where intra-person visual differences are often larger than inter-person similarities. Currently, almost all state-of-the-art ReID methods are based on deep learning. Typically, such ReID methods try to embed person images into a discriminative and compact feature space, then compute the similarities between the query and gallery images. In those methods, features are represented by the last fully-connected (FC) layer.
Usually, the last FC layer extracts very high abstract features, and there is no explicit orthognality constraint on the weight vectors within the FC layer. Thus the weight vectors are easily to suffer from being highly correlated, leading to non-compact and redundant features where neurons of FC layer are correlated with each other~\cite{sun2017svdnet}.
To reduce feature redundancies which may hamper the final ranking results in ReID evaluation, Sun et al.~\cite{sun2017svdnet} propose SVDNet which uses the left unitary matrix and singular value matrix of the SVD results to replace the original weight vectors. Although the ReID performance can be consistently improved through the Restraint and Relaxation Iteration (RRI) training scheme, it might be not optimal as it breaks the end-to-end training mechanism and has to obtain the solution in an alternating variable way.

In this paper, we try to handle the problem mentioned above by utilize subspace pooling~\cite{Wei_2018_CVPR,wei2018grassmann} which transforms the CNN feature matrix into a compact orthogonal matrix consists of its principal singular vectors. Such orthogonal matrices lie on the Grassmann manifold~\cite{jayasumana2015kernel}, a Riemannian manifold where a point represents the subspace of the Euclidean space. The subspace pooling layer can be plugged into any backbone networks and trained via an end-to-end back-propagation learning method, leading to a compact and discriminative feature representations of person images shot by UAVs.

We make thorough analysis of our collected PRAI-1581 dataset, show its new features against previous datasets and conduct extensive experiments on it.
Specifically, we test 12 state-of-the-art person ReID methods and the subspace pooling method on PRAI-1581. Experimental results show that our method performs favorably against others.
Nevertheless, all the performances are far from satisfactory which demonstrate that re-identify persons in aerial imagery is very challenging and further efforts should be made in this research direction.

The contributions of this paper can be summarized from the following aspects:

\begin{enumerate}
    \item \textbf{Person ReID Dataset in Aerial Imagery.} Public datasets such as Market1501 and CUHK03 are shot under the fixed surveillance cameras. Our dataset is shot by UAVs which are flying at an altitude ranging from 20 to 60 meters. The PRAI-1581 dataset contains over 39,461 images of 1581 person identities and to our best knowledge, it is the first large person ReID dataset in aerial imagery facing to the real UAV surveillance scenarios.
    \item \textbf{Subspace Pooling for Person ReID.} We propose to utilize subspace pooling of CNN features to represent the input person images by a discriminative and compact feature representation, which achieves state-of-the-art performance on our collected dataset. The method can make the entries of the feature independent and the neurons learn de-correlated descriptors for person images.
    \item \textbf{Comprehensive Experimental Results.}
    We conduct thorough experiments on our dataset, including typical baseline, representative state-of-the-art ReID methods, and our subspace pooling method. To test the practical performance of the ReID on our dataset.

\end{enumerate}

The rest of the paper is organized as follows: Section~\ref{sec_rw} reviews the related works of our paper.
Section~\ref{sec_dataset} describes our PRAI dataset in detail.
In Section~\ref{sec_approach} we elaborate the approach of subspace pooling with SVD. Section~\ref{sec_exp} gives the experimental results.
In Section\ref{sec_conclusion} we draw the conclusion of the paper.



\section{Related works}\label{sec_rw}

In this section, we review the related works from the following three aspects: aerial datasets for vision tasks, person ReID methods and compact feature learning.

\subsection{Aerial Datasets for Vision Tasks}
In recent years, more and more aerial datasets have emerged,
featuring at scale diversity, perspective specificity, multi-directional and high background complexity.
DOTA~\cite{xia2018dota} is an object detection dataset in aerial images which are
mainly collected from the Google Earth, satellite JL-1, and satellite GF-2 of the China Centre for Resources Satellite Data and Application. It contains 0.4 million annotated object instances within 16 categories.
NWPU VHR~\cite{cheng2016learning} is an aerospace remote sensing dataset including 650 targets and 150 scene images, and the main purpose of the dataset is serving as an detection benchmark, too.
UAV123~\cite{mueller2016benchmark} is a video dataset taken by UAVs and aimed to serve as a target tracking dataset.
VisDrone~\cite{zhu2018vision} is a visual object detection and tracking dataset, serving as a benchmark dataset for a challenge named ``Vision Meets Drone: A Challenge'' organized as a workshop in ECCV and ICCV.
AVI~\cite{singh2018eye} dataset is collected to deal with the task of violent individuals identification, which is composed of 2000 aerial images and annotated with 14 key-points for each person.
As for the field of Person ReID, the datasets commonly used are Market1501~\cite{zheng2016person}, MARS~\cite{zheng2016mars}, DukeMTMC-reID~\cite{zheng2017unlabeled} and CUHK03~\cite{li2014deepreid} etc., which are shot by the cameras with fixed emplacement with fixed angle.
MRP~\cite{layne2014investigating} is a public dataset for mobile re-identification. The purpose of the dataset is to verify the ReID algorithms on mobile platform.
Although it is taken by UAVs, the UAVs are flying at a relatively low altitude (more or less higher than normal person height) and it does not match the practical UAV surveillance scenarios.

In this paper, we collect PRAI-1581 dataset for person ReID task facing to the real UAV surveillance scenarios.
Our dataset  provides images from both all sorts of shooting angles and diverse resolutions duo to the cameras and the UAV platform that can be controlled flexibly, which greatly meets the research needs of person ReID in aerial imagery. Meanwhile, it is the first aerial person ReID dataset facing to real surveillance scenarios to our best knowledge.

\subsection{Person ReID Methods}
Person ReID has been attracting a lot of research efforts in the community. Traditionally, it has been addressed to deal with from two aspects: the feature representation and metric learning problem, while current state-of-the-art ReID methods are all based on deep learning techniques~\cite{Zheng2016Zero,Alam2017A,ahmed2015improved,liu2015set,yi2014deep,liu2018part,xiang2018homocentric,chen2018person}, which jointly optimizes the two phases together.
Zhou \emph{et al.}~\cite{zhou2017large} proposed a set to set distance to learn discriminative and stable feature representations and to effectively find out the matched target to the probe object among various candidates.
Ge \emph{et al.}~\cite{Ge2018FD} proposed to utilize pose information and adversarial learning to distill identity-related and pose-unrelated feature representation.

In addition, unsupervised and cross-modal ReID methods, also have achieved promising performance in recent years.
Li \emph{et al.}~\cite{li2018unsupervised} proposed an unsupervised deep ReID approach, which can incrementally discovered and exploited the underlying discriminative information for ReID.
Yu \emph{et al.}~\cite{Yu2019Unsupervised} introduced a deep model for the multi-label learning for unsupervised ReID, which learns a discriminative embedding for the unlabeled target domain through the soft multi-label guided hard negative mining.
Shen \emph{et al.}~\cite{shen2018person} proposed SGGNN and utilized the relationship between probe-gallery pairs to provide more consistent information.
Wei \emph{et al.}~\cite{wei2018person} adopted GAN to bridge the domain gap between different person ReID dataset.
Li \emph{et al.}~\cite{li2018harmonious} utilized a new spatio-temporal attention module based on person video to allow useful information to be extracted.
In this work, we study the effect of subspace pooling on person ReID task, which takes full advantage of the matrix orthogonality to obtain discriminative and invariant features through an end-to-end learning way.

\subsection{Compact Feature Learning}

Lots of research works have been focusing on learning compact and non-redundant feature descriptors which have many good merits~\cite{sun2018dissecting,Wei_2018_CVPR,xie2017all,sun2017svdnet,chan2015pcanet,wei2018grassmann}.
Chan \emph{et al.}~\cite{chan2015pcanet} proposed a very simple deep learning framework called PCANet for image classification, which learned orthogonal projection to produce the filters.
Xie \emph{et al.}~\cite{xie2017all} utilized the regularization effect of orthogonality to improve the classification accuracy.
Sun \emph{et al.}~\cite{sun2018dissecting} quantitatively analyzed the influence of features on person ReID accuracy and found that the associations between different features also impacted the results.
Wei \emph{et al.}~\cite{Wei_2018_CVPR,wei2018grassmann} proposed subspace pooling layer to learn discriminative and invariant features.
Sun \emph{et al.}~\cite{sun2017svdnet} exploited the SVDNet to generate de-correlated descriptors suitable for person retrieval, which directly benefit the feature learning process.
Although the ReID performance can be consistently improved by SVDNet, it might be not optimal as it breaks the end-to-end training procedure and has to obtain the solution in the alternating variable way.
In this work, different from SVDNet, we incorporate subspace pooling layer to transform convolution feature maps into compact feature descriptors through an end-to-end learning method.

\section{DATASET}\label{sec_dataset}

In this section, we elaborate the features of our PRAI-1581 dataset, including the collection and annotation process in detail.

\subsection{Dataset Collection}
We use two UAVs to shoot videos in two adjacent areas where there are non-overlap visible areas.
PRAI-1581 contains pairing data from two flights which are in unconstrained and heavily crowded outdoor environments.
In order to acquire more sufficient video including diverse viewpoints and backgrounds, we adopted the hovering, cruising, and rotating sport models in the process of controlling UAV.
We collect about 60 pairs of videos  and then sample the video clips at the rate of one frame for every second in each clip.
Finally, we obtain 39,461 person images including 1581 person instances.

\subsection{Annotation}
We utilize an annotation tool to label the captured video images.
The whole annotating process is divided into three steps.
The first step is object bounding box annotation, where all visible persons are manually marked with their positions in the images. With the bounding box positions, our dataset can also served as an aerial person detection and tracking benchmark dataset.
The annotated product of this step is the smallest rectangle bounding box containing the complete person, which can be stored automatically.

The second step is cross-camera person matching and person ID Assignment, which is the most time-consuming step in the entire process.
The same person appearing in two video clips need to be grouped together by manually searching and grouping. The same person instances are assigned with an unique number, ranging from $0\sim 1580$.
Then the annotation files which contain the position of person instances and their ID numbers are generated automatically for each group.
Finally, we crop the person instances from the original images to get the aerial person ReID dataset according to the information provided by the generated annotation files.

\begin{figure*}
  \centering
  \includegraphics[width=0.9\linewidth]{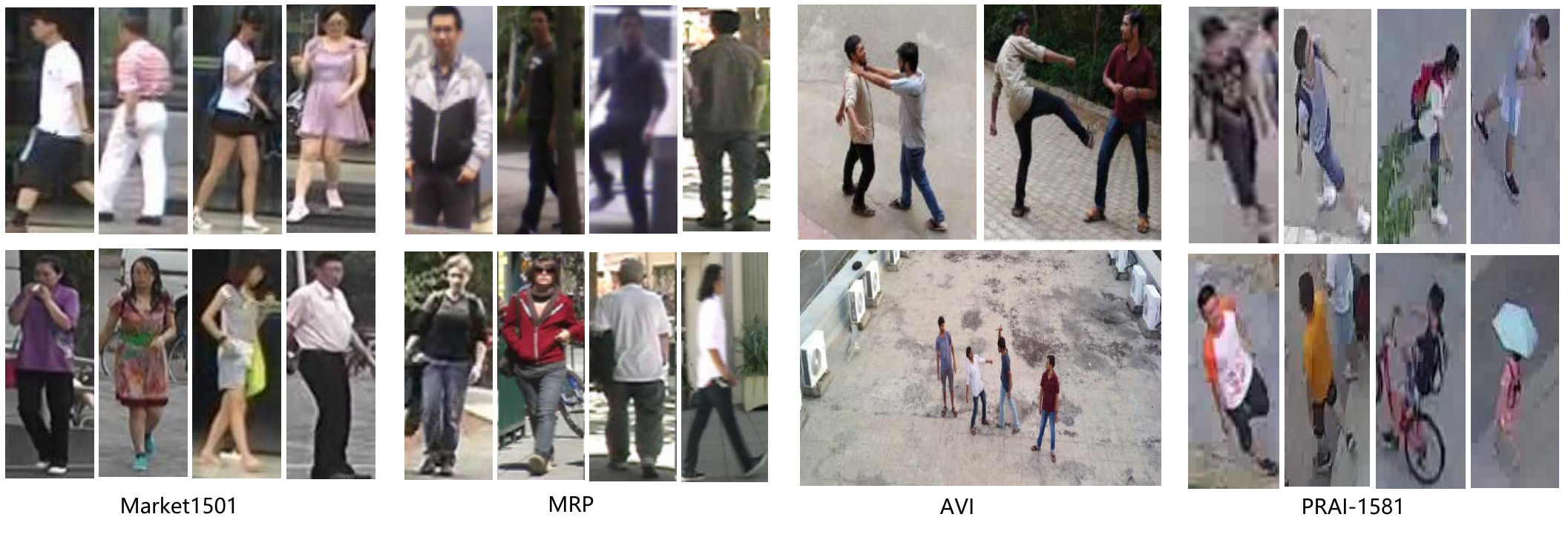}
  \caption{Example images of Market-1501, MRP, AVI and our PRAI-1581.}\label{dataset_comp}
\end{figure*}

In addition, to more comprehensively illustrate our dataset, we compare our PRAI-1581 dataset with the Market1501~\cite{zheng2016person}, MRP~\cite{layne2014investigating} and AVI~\cite{singh2018eye} dataset from six aspects, including No. of images, IDs, and UAVs, the camera setting platform, height, and the situation of the camera.
As can be seen from Table~\ref{tab_dataset},  Market-1501~\cite{zheng2016person} is a commonly used ReID dataset under traditional surveillance settings. MRP~\cite{layne2014investigating} is the first ReID dataset captured by UAVs while its purpose is to verify the ReID algorithms on mobile platform, so the settings of the flight are not facing to the practical UAV surveillance scenarios. AVI~\cite{singh2018eye} is a dataset for violent activity identification which is also captured by UAVs. Compared to the above described datasets, our PRAI-1581 is the first person ReID dataset facing to the practical UAV surveillance scenarios.
To visually compare the four datasets, we show some example images of each dataset in Figure~\ref{dataset_comp}.

\begin{table*}[htbp]
\centering
\fontsize{10}{15}\selectfont
\caption{Detailed settings comparison between our PRAI-1581 and other related datasets.}
\label{tab_dataset}
   \begin{tabular}{|l|c|c|c|c|c|}
    \hline
     &Market1501~\cite{zheng2016person}&MRP(Dataset1)~\cite{layne2014investigating}&MRP(Dataset2)~\cite{layne2014investigating}&AVI~\cite{singh2018eye}&PRAI-1581\cr
    \hline
     {\bf \#Images}&32668 &1501&4096&10863& 39461 \cr
    \hline
     {\bf \#ID}&1501 &23&28&5124& 1581 \cr
    \hline
     {\bf Platform}&fixed emplacement& UAV&UAV &UAV& UAV \cr
    \hline
    {\bf Height}&$\times(<10m)$ &$\times(< 10m)$&$\times(<10m)$&2$\sim$8m&20$\sim$60m \cr
    \hline
    {\bf Camera}&fixed& mobile&mobile&mobile& mobile \cr
    \hline
    {\bf \#UAVs}&0 &3&6&$\times$& 2 \cr
    \hline
    \end{tabular}
\end{table*}

\begin{figure}
  \center
  \includegraphics[width=0.9\linewidth]{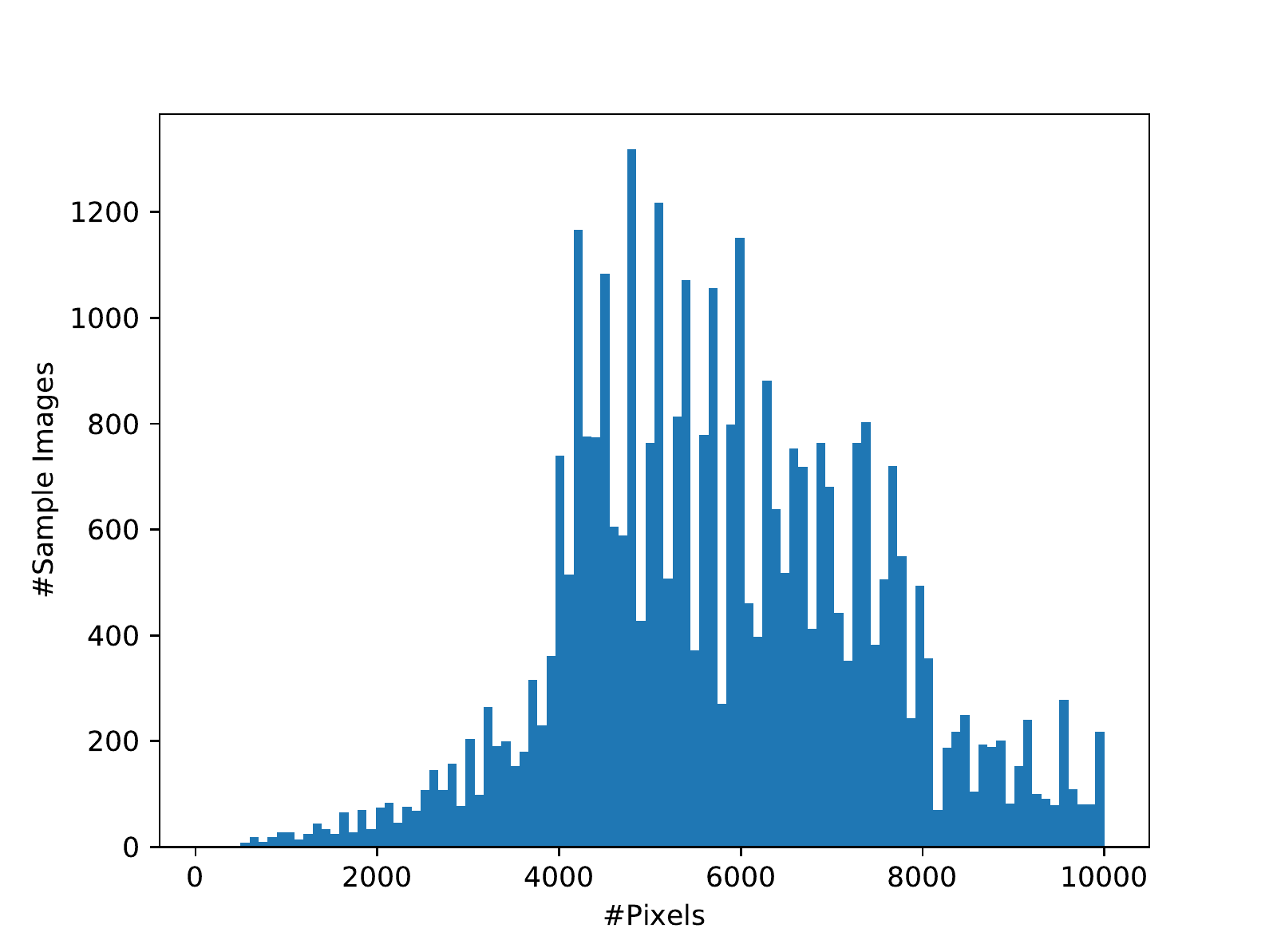}\\
  \caption{The statistical distribution over resolution (number of pixels) for our dataset. The horizontal axis represents the size of resolution (number of pixels); The vertical axis represents the number of sample images.}\label{dataset-fenbu}
\end{figure}

\subsection{Features of Our Dataset}

\begin{figure}
  \center
  \includegraphics[width=0.8\linewidth]{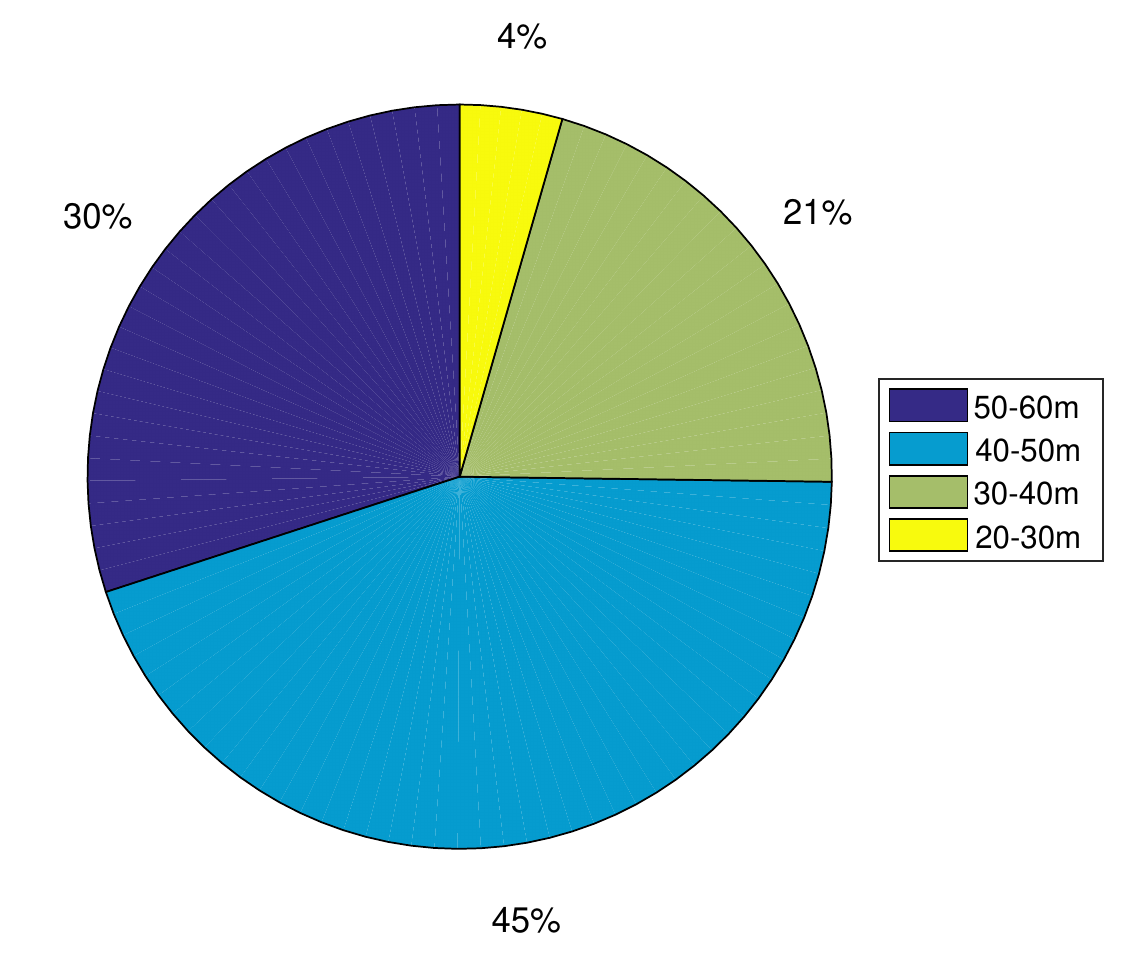}\\
  \caption{The sample distribution over UAV flying altitude in our dataset.}\label{height_distribution}
\end{figure}

\begin{figure}
  \center
  \includegraphics[width=0.99\linewidth]{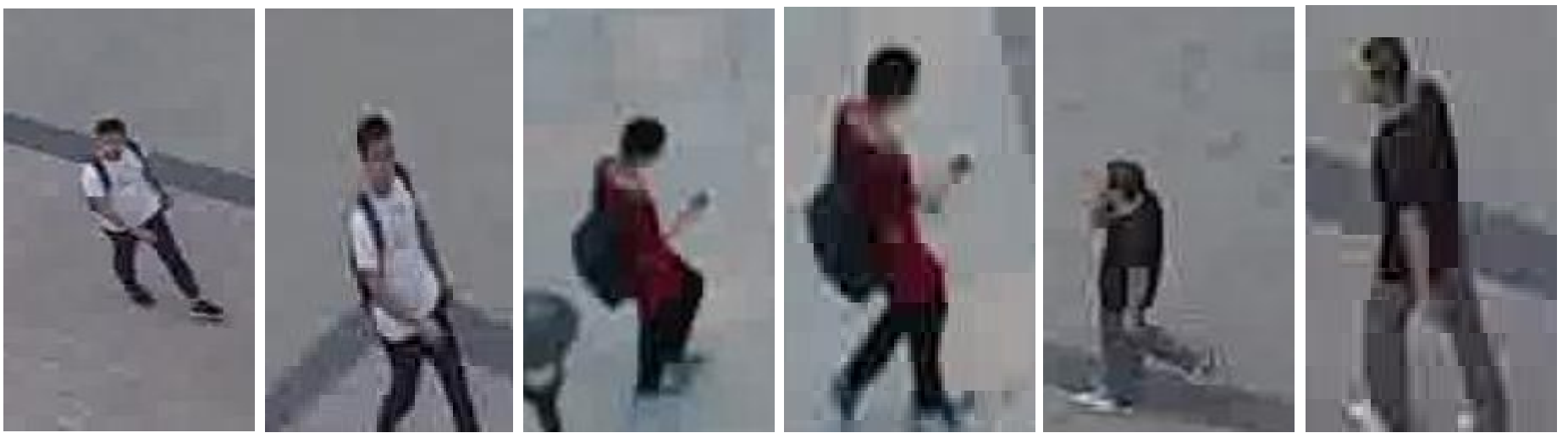}\\
  \caption{Multiple scale images under  UAVs flying at different altitudes: the first two images are from the same person, the next two images are from the second person and the last two images are from another person.}\label{height}
\end{figure}

{\flushleft\textbf{Captured by UAVs flying at rich altitudes.}}
To increase the diversity of the images, the UAVs are controlled to fly at different altitudes, ranging from 20 to 60 meters across an outdoor environment.
Figure~\ref{height} shows person images which are captured at different altitude, and it can be seen that
our dataset has rich scale diversity, which is much more challenging than single scale persons in traditionally person ReID task with fixed camera emplacement.
Figure~\ref{height_distribution} shows the sample distribution over the UAV flying height for our dataset.

\begin{figure}
  \center
  \includegraphics[width=0.99\linewidth]{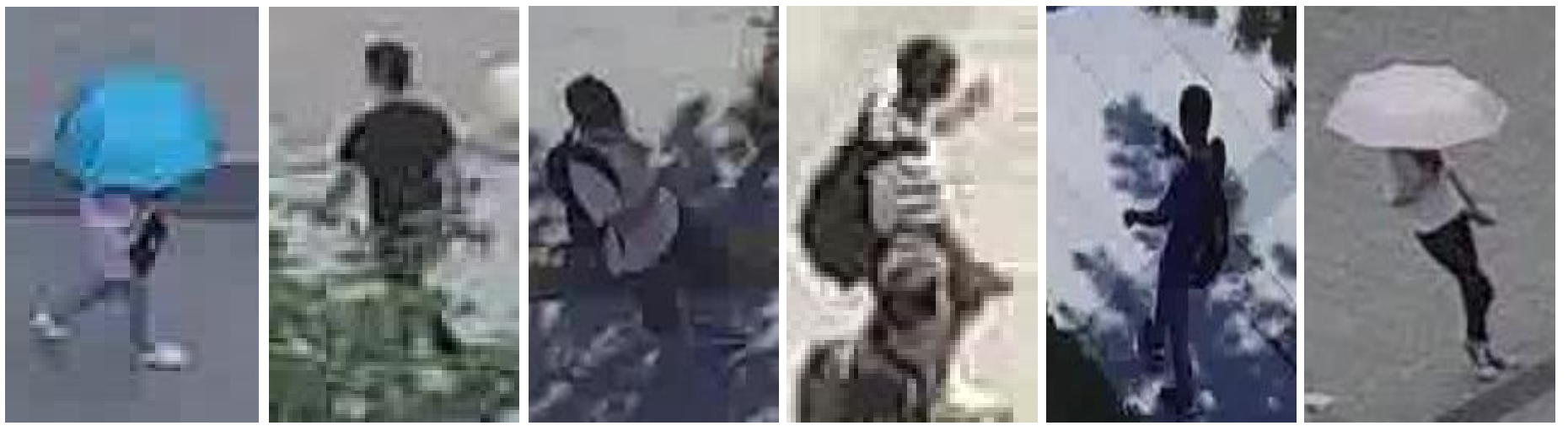}\\
  \caption{Images in diverse occlusion situations in our PRAI-1581 dataset.}\label{occlusion}
\end{figure}

{\flushleft\textbf{Low resolution.}}
As shown in Figure~\ref{occlusion}, the size of each holistic image is $4K\times 2K$, while the resolution of the captured person is very low, ranging from about 30 to 150 pixels, which is far lower than traditional ReID person images.
Low resolution is an especially challenging factor in aerial person ReID task.
The statistical distribution over resolution (number of pixels) for our dataset is shown in Figure~\ref{dataset-fenbu}. It can be clearly seen that the average resolution is far lower than that in Market-1501.

\begin{figure}
  \center
  \includegraphics[width=0.99\linewidth]{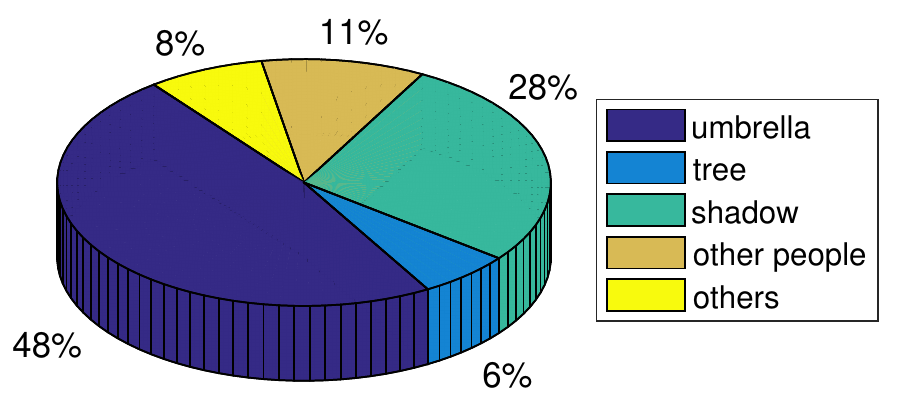}\\
  \caption{The sample distribution over occlusion types in our dataset.}\label{occlusion_types}
\end{figure}

{\flushleft\textbf{Occlusions.}}
Severe and multi-type occlusions is another challenging factor in our dataset. Figure~\ref{occlusion} shows some examples with different occlusions. In our dataset, there are about $20\%$ sample images are severely occluded typically by umbrellas, trees, shadows, and other persons etc.. Figure~\ref{occlusion_types} shows the sample distribution over occlusion types in our dataset.

\begin{figure}
  \center
  \includegraphics[width=0.99\linewidth]{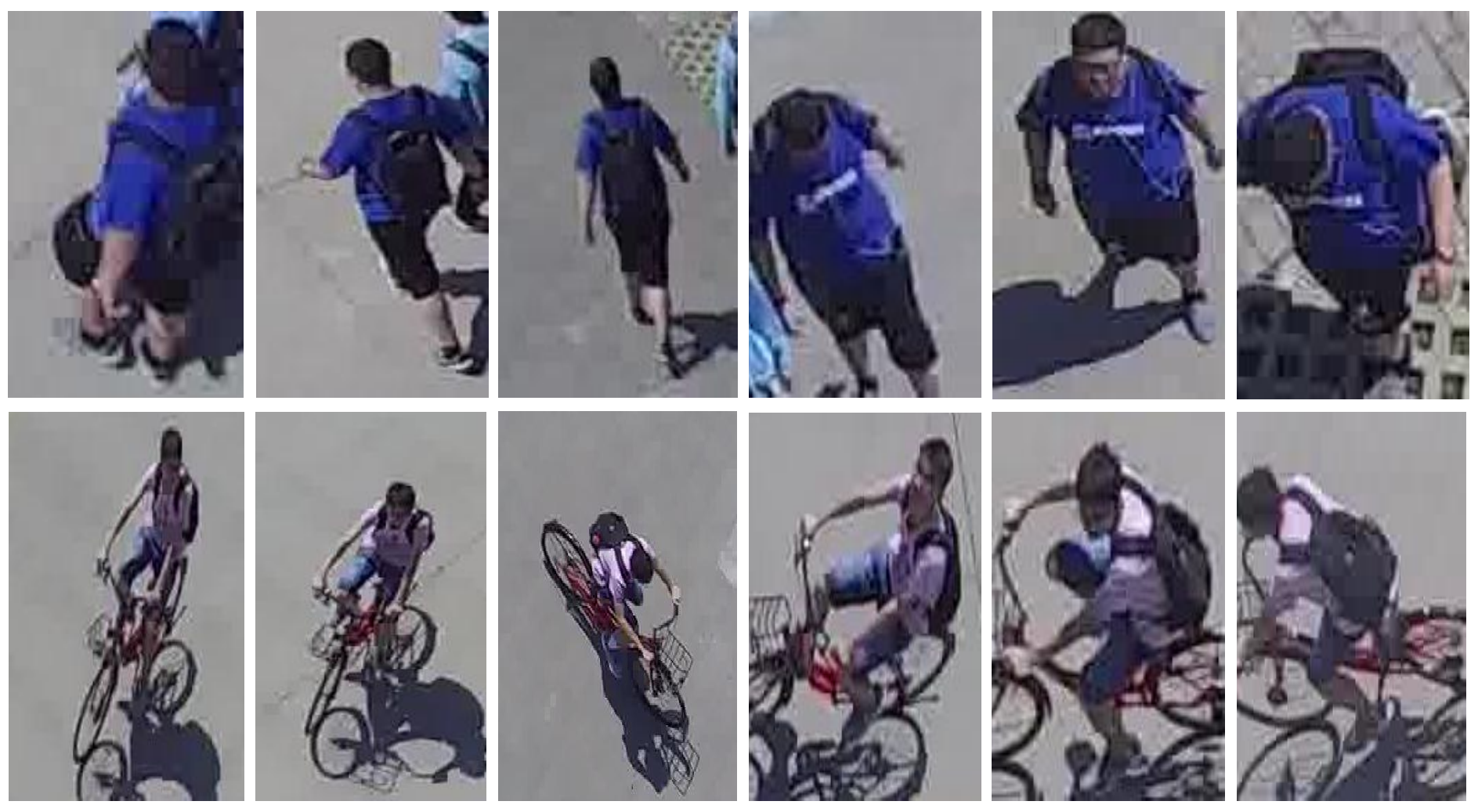}\\
  \caption{Images captured by UAVs exhibits various perspectives and poses. Images in each row have the same person ID.}\label{view}
\end{figure}

{\flushleft\textbf{Diverse views and poses.}}
Figure~\ref{view} shows some images for one same person captured by UAVs under different perspectives.
Our PRAI-1581 dataset includes diverse views and poses, including various profile views and top views, as UAVs can dynamically adapt the viewing positions and directions, while without being constrained by the fixed locations. So our dataset can meet the research need of recognition or matching related tasks for diverse angles, views and poses under complex conditions.

Due to the rich characteristics described above, our dataset can not only be served as a benchmark dataset for aerial person ReID facing to practical UAV surveillance scenarios, but also be fully tested for perception problem at a distance while aiming at building scale, pose, view (especially across profile view and top view), and occlusion invariant feature representations.

\section{Approach}\label{sec_approach}
\begin{figure}
  \center
  \includegraphics[width=0.99\linewidth]{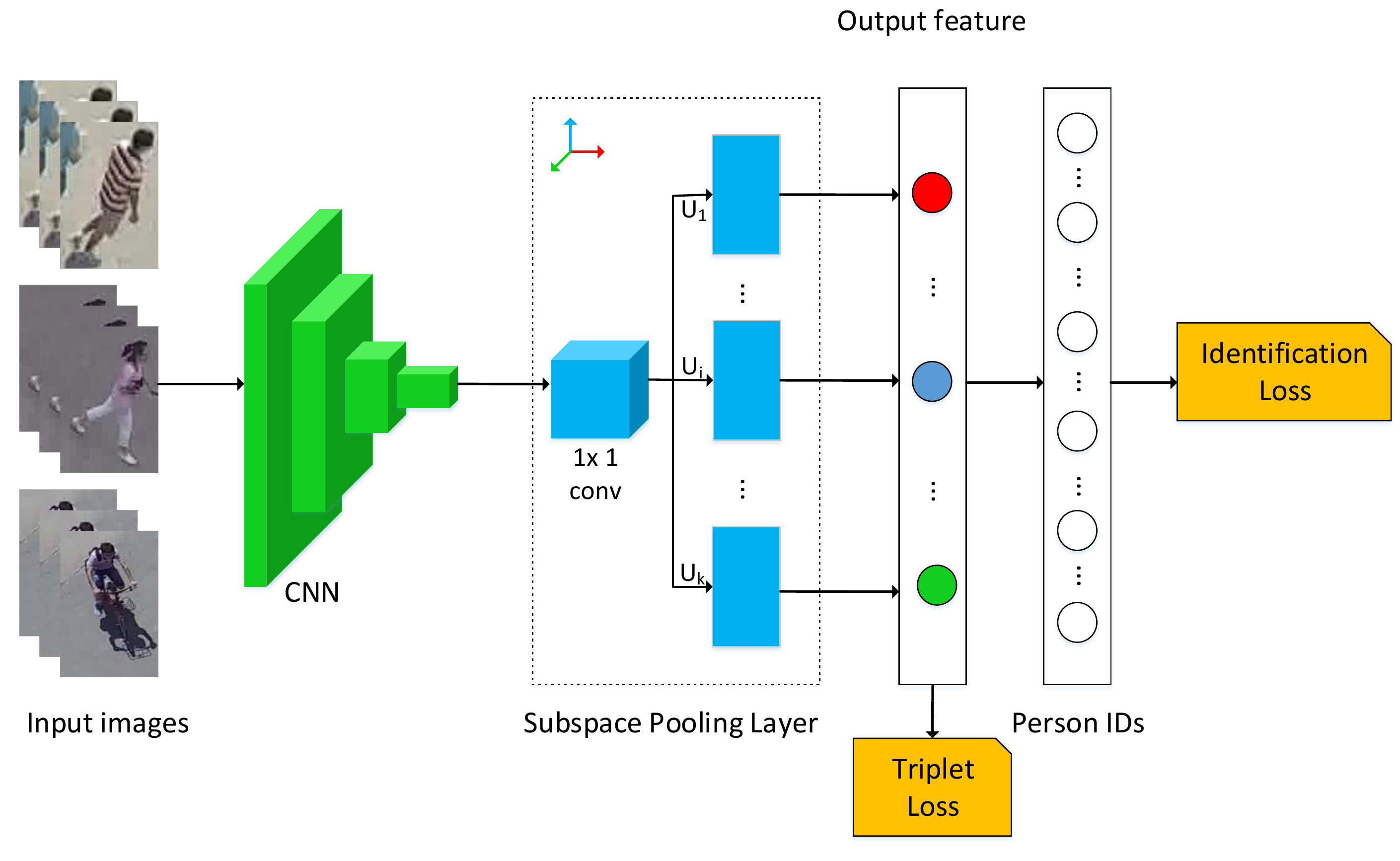}\\
  \caption{The framework of the proposed approach, a subspace pooling layer is integrated with a backbone CNN model.}\label{net}
\end{figure}

In this section, we elaborate the ReID algorithm based on subspace pooling, which achieved state-of-the-art performance on our dataset compared with many baseline methods. The framework of the proposed approach is shown in Figure ~\ref{net}.
The subspace pooling layer is integrated with a backbone CNN model, and the architecture is fully differentiable so it can be optimized via standard back-propagation algorithm.
Firstly, to emphasize the difference with the SVDNet~\cite{sun2017svdnet}, we briefly revisit the SVDNet method.

\subsection{SVDNet Revisit}
SVDNet makes an assumption that the featuring person descriptor in the last fully connected layer remains redundant and correlated. To eliminate the correlation, they adopt the SVD technique to decompose the weight matrix of the last fully-connected layer. As the weight matrix can be viewed as the projection basis, replacing the original weight matrix by using the left singular vectors weighted by the singular values ensures the orthogonality of the projection basis, thus leading to non-correlated feature descriptors in the fully-connected layer.
Through the Restraint and Relaxation Iterative (RRI) training scheme, the orthogonality constraint can be integrated with the CNN training process.
Note that the RRI training scheme is not an end-to-end way, CNN training and SVD decomposition are iteratively performed until convergence.
On the other hand, we utilize subspace pooling to project the original CNN features into an orthogonal matrix while keeping the whole framework end-to-end trainable.

\subsection{Subspace Pooling}
In order to achieve the goal of reducing the feature redundancy and obtaining compact feature descriptors,
we utilize the subspace pooling layer~\cite{wei2018grassmann} combined with the backbone CNN to get the final feature representation.
Specifically, denote $\mathbf{A}\in \mathbb{R}^{c\times hw}$ as the last convolution feature maps of the backbone CNN model, where $c$, $h$, $w$ represents channel, height and width of the feature maps respectively. Each row of $\mathbf{A}$ corresponds to one feature map enrolled into a $1$-D vector. Each column of $\mathbf{A}$ represents one $c$-D local feature vector at a specified spatial location.
Generally, the SVD result of the CNN features can be formulated as
\begin{equation}
\begin{aligned}\label{eqn_bat}
    \mathbf{A} = \mathbf{U} \mathbf{S} {\mathbf{V}}^T = \sum_{i=1}^{c}{\sigma_i {\mathbf{u}}_i {\mathbf{v}}_i^T}
\end{aligned}
\end{equation}
Without loss of generality, we assume that the singular values and vectors are arranged as the descending order of $\sigma$.
In many cases, the sum of the top 10\% or even 1\% of the singular values accounts for 99\% of the sum of all the singular values.
Thus we can approximately describe the matrix with the largest $k$ ($k \ll c$) singular values and the corresponding left singular vectors and right singular vectors. In addition, it is easy to prove that $\mathbf{A}_k = \sum_{i=1}^{k} {\sigma_i \mathbf{u}_i \mathbf{v}_i^T}$, is the best rank $k$ matrix approximations under the metric of Frobenius norm~\cite{wei2018grassmann}.

The output of the subspace pooling layer can be formulated as,
\begin{equation}
\begin{aligned}\label{eqn_subspace_pooling}
    f(\mathbf{A}) = \mathbf{U}_k = [\mathbf{u}_1,\mathbf{u}_2,\cdots,\mathbf{u}_k]
\end{aligned}
\end{equation}
That is, the subspace pooling layer transforms the input convolution feature maps into its $k$ left singular vectors corresponding to the largest $k$ singular values.
The obtained orthogonal matrix geometrically supports a subspace of the Euclidean space.

\subsection{Loss Function}
In practice, to further improve the efficiency, a dimension reduction layer is introduced before we feed the convolution feature maps into the subspace pooling layer. Specifically, 1$\times$1 conv layer is utilized to reduce the feature dimensionality. After the subspace pooling layer, a linear classification layer is incorporated to project the  feature vector to the person IDs.

We use two types of loss functions, softmax with cross entropy loss and triplet loss.
The softmax with cross entropy loss can be formulated as follows:
\begin{equation}
\begin{aligned}\label{eqn_bat2}
    \mathcal{L}_{CroEn} = -\frac{1}{N}\sum_{i=1}^{N}\sum_{t=1}^{T}y_{i,t}\log{p_{i,t}}
\end{aligned}
\end{equation}
Where $N$ is the number of samples, $T$ is the number of target person IDs, $y$ is the ground truth label, $p$ is the predicted label after the softmax activation function.

To take full advantage of the hard triplet sampling mechanism, the batch hard triplet loss is adopted, which can be formulated as,
\begin{equation}
\begin{aligned}\label{eqn_bat3}
    \mathcal{L}_{Tri} =  \sum_{i = 1}^P \sum_{a = 1}^K
    \max \big(0, {D_{a,p}^{*}} - {D_{a,n}^{*}} + m \big)
\end{aligned}
\end{equation}
Where $P \times K$ is the mini-batch size, sampling with P person IDs and K instances for each ID.
$D_{a,p}^{*}$ is the largest distance among those between anchor sample and positive samples, and
$D_{a,n}^{*}$ is the smallest distance among those between anchor sample and negative samples. $m$ is a hyper-parameter which denotes the margin in triplet loss.
In order to verify the effectiveness of subspace pooling layer, we conduct extensive experiments in Section~\ref{sec_exp} on our collected dataset with many baselines and the proposed method achieves state-of-the-art performance compared with the baseline methods.
Detailed results are shown in Section~\ref{sec_exp}.

\subsection{Connections Between Subspace Pooling  and SVDNet.}
Although the subspace pooling layer and SVDNet both are benefited from the orthogonal projection or SVD technique, they differs in two major aspects.
Firstly, SVDNet performs SVD on the weight vectors within the last FC layer, while subspace pooling performs SVD on the CNN features maps which are projected 
into a Riemannian manifold~\cite{jayasumana2015kernel}.
Secondly, their optimization schemes are totally different. Restraint and Relaxation Iteration (RRI) is the key procedure in training SVDNet, which is involved with iterative SVD decomposition and CNN training steps.
However, The parameters of the whole model in subspace pooling method are learned in an end-to-end manner, which makes the training method more efficient and effective.
Thus, although both subspace pooling and SVDNet establish a connection between CNN and SVD, their network models and operating mechanisms are totally different. In the experiments, we show that the subspace pooling method outperforms the SVDNet by a large margin on our dataset.

\section{Experiments}\label{sec_exp}

In this section, we empirically evaluate many baseline methods and representative state-of-the-art person ReID algorithms on our collected PRAI-1581 datasets, then the proposed subspace pooling method is reported and compared. Firstly, for fair comparison, we split the dataset into training and test set for only once and fix the partition of dataset for all the experiments. Then, we introduce the implementation details and evaluation metrics in Section~\ref{sec_detailandmetric}. The results of the baseline and state-of-the-art methods are reported in Section~\ref{sec_baselines}. And the proposed subspace pooling method is evaluated in Section~\ref{sec_sppooling}. Lastly, we make some ablation studies in Section~\ref{sec_ablas}

\subsection{Dataset Partition}\label{sec_datasplit}
Our PRAI-1581 dataset contains a total of 39,461 images for 1581  individual identities.
We randomly divide the dataset into training and test set for only once and keep the partition fixed in all the experiments for fair comparison.
The ratio of training and testing set is roughly set to 1:1.
The training set includes 19,523 images of 782 identities.
And the rest of the dataset is used as the test set with 19,938 images of 799 identities. To be more specific,
the numbers of query and gallery images are 4680 and 15,258 respectively.

\subsection{Implementation Details and Evaluation Metrics}\label{sec_detailandmetric}

In our experiments, we use ResNet-50 as the backbone CNN model as it is a very popular model for the ReID task.
The parameters of the backbone model are pre-trained on the ImageNet-2012 dataset.
The batchsize is set to $32\times 4=128$, sampling with 32 different identities and 4 instances per identity in each mini-batch.
The training process takes 300 epochs in total.
The Adam optimizer with an initial learning rate of $2 \times 10^{-4}$ is used, and learning rate is decayed at the $151^{th}$ epoch according to the exponential decay rule.
The method is implemented based on the Pytorch platform and tested on a single NVIDIA 1080Ti GPU card.

For the evaluation of the experiments, we use the most common metrics in the field of person ReID, mean Average Precision (mAP) and Cumulative Matching Characteristics (CMC) at rank-1.
In addition, when evaluating the proposed subspace pooling method, we use the F-score as an additional metric to indicate its practical performance.

\subsection{Baseline and State-of-the-art ReID Methods on PRAI-1581 Dataset}\label{sec_baselines}
We report the performances of some typical methods and representative state-of-the-art person ReID methods on our dataset in Table~\ref{tab_baseandsota}. For all the baseline methods, ResNet-50 is adopted as the backbone CNN model. Firstly, we enforce the identification loss (denoted as ID) and batch-hard triplet loss~\cite{hermans2017defense} (denoted as TL) on top of the backbone model respectively. The mAP and rank-1 accuracy of the ID method is 31.47\% and 42.62\% on our dataset, while the TL method achieves a mAP of 36.49\% and a rank-1 accuracy of 47.47\%, which shows that the batch-hard triplet loss performs better on our dataset for aerial person ReID.
\begin{table}[htbp]
\centering
\fontsize{10}{15}\selectfont
\caption{Results of Baseline and State-Of-The-Art ReID Methods on PRAI-1581 Dataset. ID denotes Identification Loss, TL denotes batch hard Triplet Loss, STL denotes Strong Triplet Loss.}
\label{tab_baseandsota}
   \begin{tabular}{c|c|c}
    \hline
     method&mAP&rank-1\cr
    \hline
     \hline
     ID& 31.47& 42.62 \cr
    \hline
     TL~\cite{hermans2017defense}& 36.49& 47.47 \cr
     \hline
     PCB~\cite{sun2018beyond}& 37.15& 47.47 \cr
     \hline
     STL~\cite{sun2018beyond}& 37.13& 47.49 \cr
     \hline
     \hline
     SVDNet~\cite{sun2017svdnet}& 36.70& 46.10 \cr
    \hline
     AlignedReID~\cite{zhang2017alignedreid}& 37.64& 48.54 \cr
     \hline
     PCB+RPP~\cite{sun2018beyond}& 38.45& 48.07 \cr
    \hline
     MBC~\cite{ustinova2017multi}& 22.83& 30.05 \cr
     \hline
     2Stream~\cite{zheng2018discriminatively}& 37.02& 47.79 \cr
     \hline
     Part-align~\cite{zhao2017deeply}& 32.86& 43.14 \cr
     \hline
     DSR~\cite{he2018deep}& 39.14& 51.09 \cr
     \hline
     IDE~\cite{zhong2018camera}& 32.90& 43.90 \cr
     \hline
     DCGAN~\cite{zheng2017unlabeled}& 28.82& 38.93 \cr
     \hline
     Deep Embedding~\cite{ahmed2015improved}& 14.73& 21.36 \cr
     \hline
     MGN~\cite{wang2018learning}& 40.86& 49.64 \cr
     \hline
     OSNET~\cite{zhou2019omni}& 42.10& 54.40 \cr
     \hline
    \end{tabular}
\end{table}

Then, we test the Part-based Convolutional Baseline (PCB) which is proposed in~\cite{sun2018beyond} as a strong person ReID baseline.
Note that the PCB method conducts uniform partition on the convolution feature maps for learning part-level features and it achieves a mAP of 37.15\% and a rank-1 accuracy of 47.47\%, which outperforms the naive ID method by +5.68\% on mAP and +4.85\% on rank-1. Similarly, a part based strong triplet loss (STL) method achieves comparable performance on our dataset, with mAP of 37.13\% and rank-1 accuracy of 47.49\%.

Lastly, we report representative state-of-the-art person ReID algorithms SVDNet~\cite{sun2017svdnet}, AlignedReID~\cite{zhang2017alignedreid}, Refined Part Pooling (RPP)~\cite{sun2018beyond}, 
MBC~\cite{ustinova2017multi}, 2Stream~\cite{zheng2018discriminatively}, Part-align~\cite{zhao2017deeply}, DSR~\cite{he2018deep}, IDE~\cite{zhong2018camera}, DCGAN~\cite{zheng2017unlabeled}, Deep Embedding~\cite{ahmed2015improved},
MGN~\cite{wang2018learning}, OSNET~\cite{zhou2019omni} methods.
The SVDNet also incorporates the SVD technique to reduce the redundancy of the person features and achieves a mAP of 36.7\% and a rank-1 accuracy of 46.1\%. AlignedReID learns to align the part features during the training process and it achieves a mAP of 37.64\% and a rank-1 accuracy of 48.54\%. Based on the PCB, RPP re-assigns the outliers in the evenly partitioned initial part to enhance the within part consistency and achieves a mAP of 38.45\% and a rank-1 accuracy of 48.07\%. 
OSNET~\cite{zhou2019omni} and MGN~\cite{wang2018learning} which fuse multi-scale or multi-granularity global and part features, achieved state-of-the-art performance on our dataset.
Note that although the part-based methods performs much better than the plain counterpart baselines for ReID in traditional surveillance cameras, they are not as effective as on our dataset probably because the diverse angles and views in aerial person images makes that the upright assumption of person images in traditional surveillance cameras can not hold, thus leading to bad part alignment results.

\subsection{The Subspace Pooling Method on PRAI-1581 Dataset}\label{sec_sppooling}
We then compare the proposed subspace pooling (SP) method with baselines on our PRAI-1581 dataset and the results are shown in Table~\ref{tab_svd}.
It can be clearly seen that SP combined with ID method can improve the mAP from 31.47\% to 37.88\%, with 6.41\% gains and improve the rank-1 accuracy from 42.46\% to 48.33\%, with 5.87\% gains.
While the SP combined with TL method can improve the mAP from 36.49\% to 39.58\%, with 3.09\% gains and improve the rank-1 accuracy from 47.47\% to 49.79\%, with 2.32\% gains. Note that the ``SP+TL'' method achieves state-of-the-art performance on our dataset, as it not only outperforms the baseline methods by a large margin, but also outperforms state-of-the-art person ReID algorithms such as  SVDNet~\cite{sun2017svdnet}, AlignedReID~\cite{zhang2017alignedreid} and ``PCB+RPP''~\cite{sun2018beyond}.

\begin{table}
  \centering
  \fontsize{9}{15}\selectfont
  \caption{Results of the proposed subspace pooling method on PRAI-1581 dataset.  TL denotes Triplet Loss, ID denotes Identification Loss, SP denotes subspace pooling.}
  \label{tab_svd}
    \begin{tabular}{c|c|c|c|c|c|c}
    \hline
    \multirow{2}{*}{method}&
    \multicolumn{3}{c|}{Single-query}&\multicolumn{3}{c}{Multi-query}\cr\cline{2-7}
    &mAP&Rank-1&F-score&mAP&Rank-1&F-score\cr
    \hline
    \hline
    ID&31.47&42.46&32.38&39.69&51.95&40.34\cr\hline
   ID+SP&37.88&48.33&38.34&43.22&54.95&43.43\cr\hline
    gains&{\bf 6.41}&{\bf 5.87}&{\bf 5.96}&{\bf 3.53}&{\bf 3.00}&{\bf 3.09}\cr\hline\hline
    TL&36.49&47.47&38.49&43.31&54.95&43.43\cr\hline
    TL+SP&39.58&49.79&39.82&44.75&56.49&44.73\cr\hline
    gains&{\bf 3.09}&{\bf 2.32}&{\bf 1.33}&{\bf 1.44}&{\bf 1.54}&{\bf 1.30}\cr\hline
    \end{tabular}
\end{table}

\begin{figure*}[!th]
  \centering
  \includegraphics[width=0.87\linewidth]{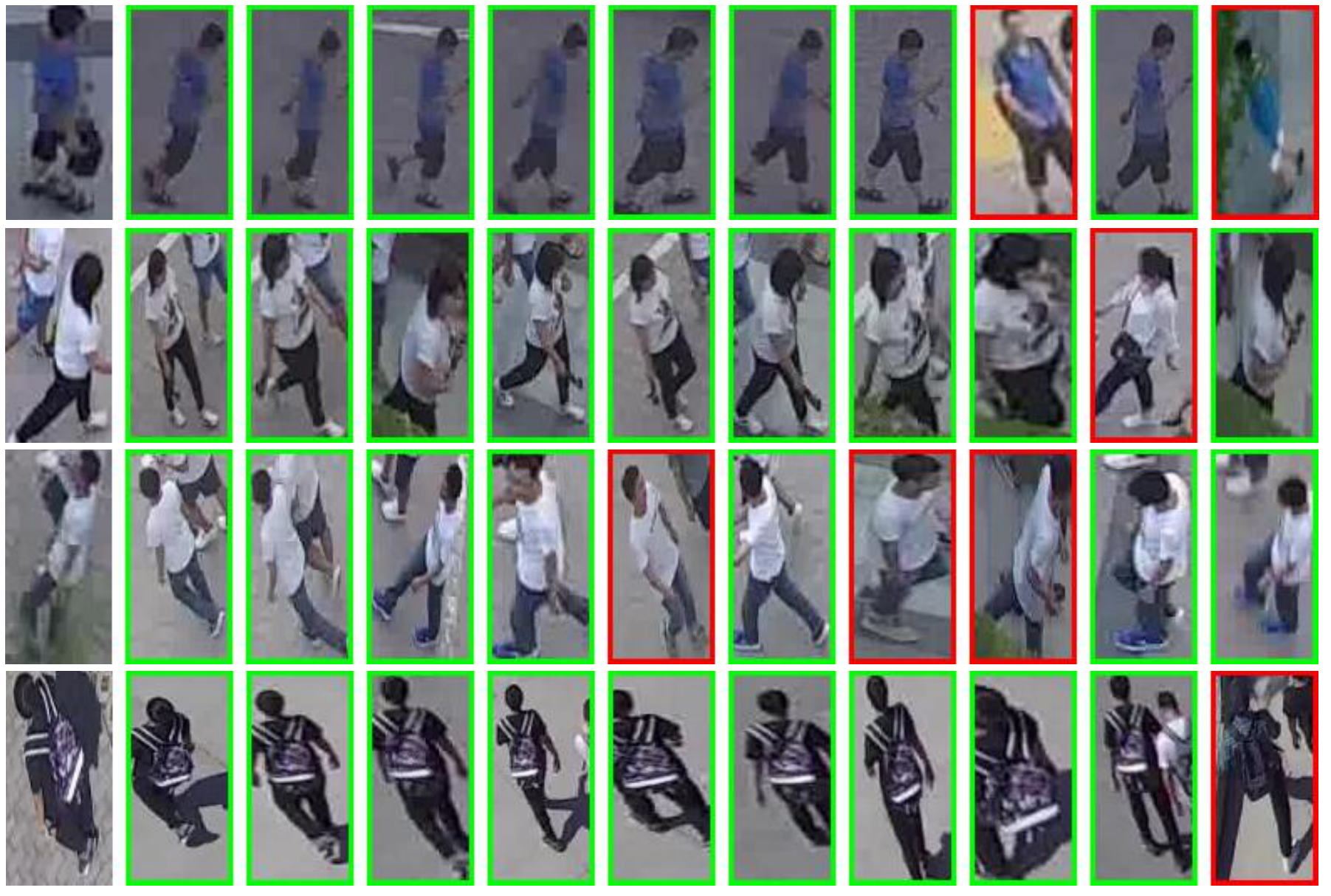}
  \caption{Successful cases of the proposed method on our dataset. Images with Green boxes indicate the true matches to the query image, and the red boxes illustrate the false matches to the query image.}\label{ranking1}
\end{figure*}

\begin{figure*}[!ht]
  \center
  \includegraphics[width=0.87\linewidth]{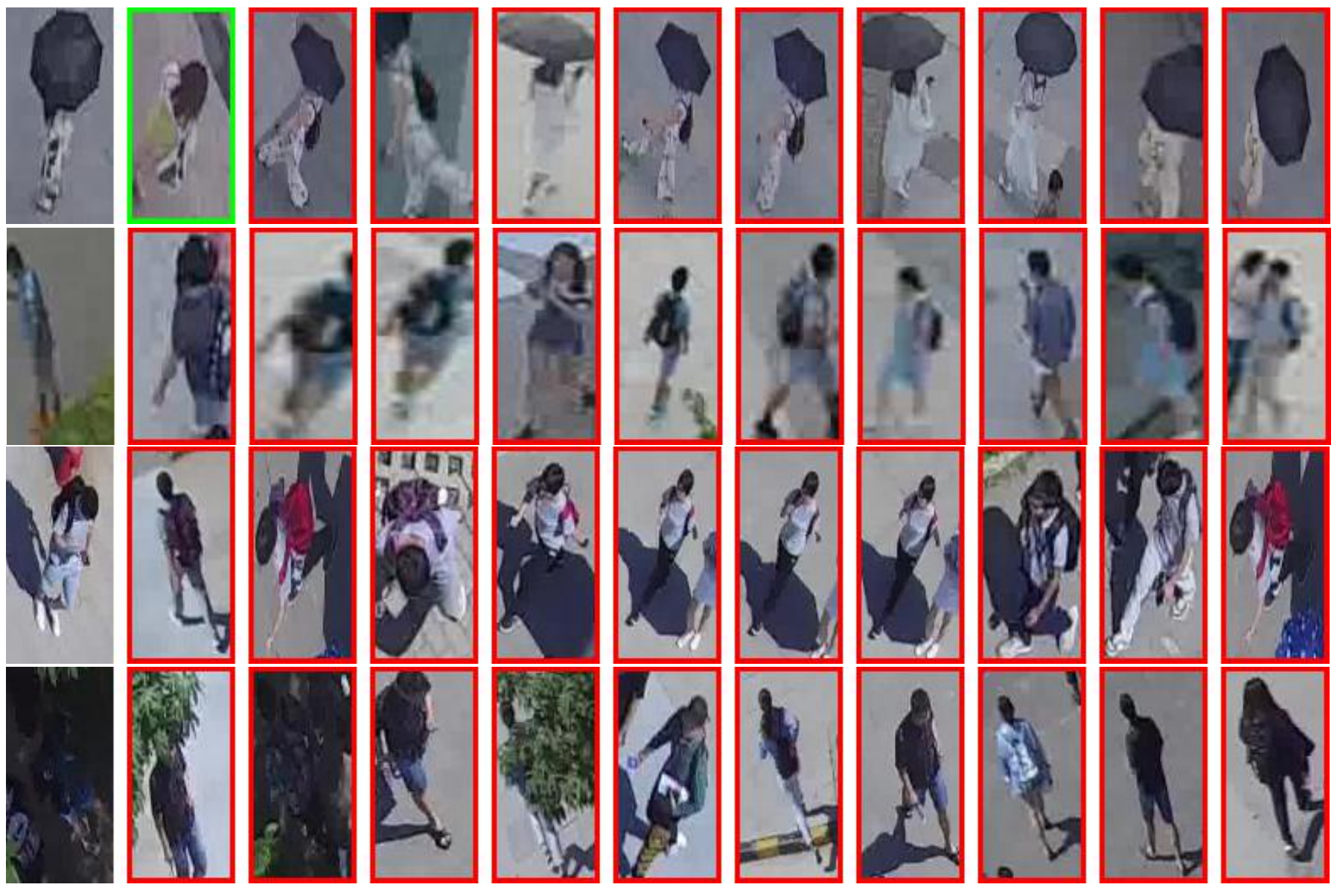}\\
  \caption{Failure cases of the proposed method on our dataset. The main reasons for the failure cases are due to occlusion (1st row), low resolution (2nd row), view angle and shadows (3rd row), illumination and shade (4th row).}\label{ranking2}
\end{figure*}

In addition, we qualitatively show some experimental results of the proposed method in Figure~\ref{ranking1} and Figure~\ref{ranking2}.
The leftmost column shows some query samples and the top-10 images in the ranking list of the gallery images are listed after the queries.
In the Figures, images with Green boxes indicate the true matches to the query image, and the red boxes illustrate the false matches to the query image.
Figure~\ref{ranking1} shows some successful cases of our method, and Figure~\ref{ranking2} gives some failure cases. It can be seen from
Figure~\ref{ranking2} that the main reasons for the failure cases are due to occlusion (1st row), low resolution (2nd row), view angle and shadows (3rd row), illumination and shade (4th row).

\subsection{Discussion}\label{sec_ablas}
{\flushleft\textbf{Dataset Partition.}}
To fully evaluate the proposed method on our dataset, we tried different partition ratios of training over test set and the results are reported in Table~\ref{tab:ratio}. Generally, more training samples can get better performance. It can also be inferred from Table~\ref{tab:ratio} that the ReID task in aerial views is much more challenging than traditional ReID with fixed camera emplacement.

\begin{table}
  \centering
  \fontsize{10}{15}\selectfont
  \caption{The results of different partition ratios of training set over test set.}
  \label{tab:ratio}
    \begin{tabular}{c|c|c|c|c}
    \hline
    \multirow{2}{*}{ratio}&
    \multicolumn{2}{c|}{Single-query}&\multicolumn{2}{c}{Multi-query}\cr\cline{2-5}
    &mAP&Rank-1&mAP&Rank-1\cr
    \hline
    \hline
    2:8&18.88&26.49&23.53&32.95\cr\hline
   3:7&25.03&33.44&31.17&42.14\cr\hline
    4:6&31.44&41.93&37.84&49.59\cr\hline
    5:5&39.58&49.79&45.23&54.74\cr\hline
    6:4&44.30&54.20&51.58&62.57\cr\hline
    7:3&50.33&60.40&58.89&69.63\cr\hline
    8:2&59.36&69.96&66.94&77.57\cr
    \hline
    \end{tabular}
\end{table}

{\flushleft\textbf{Backbone Networks.}}
We further employ different backbone networks to conduct a comprehensive performance evaluation and comparison of the proposed method on our dataset. The results are reported in Table~\ref{tab:net}. Note that when DenseNet-161 is adopted as the backbone network, it achieves better performance. However, DenseNet-161 takes much more computing and memory resources.
\begin{table}
  \centering
  \fontsize{10}{15}\selectfont
  \caption{The results of different backbone networks.}
  \label{tab:net}
    \begin{tabular}{c|c|c|c|c}
    \hline
    \multirow{2}{*}{network}&
    \multicolumn{2}{c|}{Single-query}&\multicolumn{2}{c}{Multi-query}\cr\cline{2-5}
    &mAP&Rank-1&mAP&Rank-1\cr
    \hline
    \hline
    GoogLeNet&34.86&44.94&38.80&51.32\cr\hline
    ResNet-34&35.99&47.21&41.03&52.86\cr\hline
    ResNet-50&39.58&49.79&44.75&56.49\cr\hline
    ResNext-50&37.30&48.13&43.00&54.39\cr\hline
    DenseNet-161&\bf{43.05}&\bf{54.76}&\bf{50.97}&\bf{62.41}\cr\hline
    MobileNet&33.17&44.55&38.95&51.12\cr\hline
    ShuffleNet&19.02&26.71&35.35&47.70\cr\hline
    \end{tabular}
\end{table}

{\flushleft\textbf{Results on Other Datasets.}}
To further verify the effectiveness of the proposed method, we compare the proposed subspace pooling method and its baseline on Market-1501, DukeMTMC and CUHK-03 datasets. The results of the experiments are all based on ResNet-50 backbone network.
As can be seen from Table~\ref{other_datasets}, the baseline method achieves 71.44\% mAP and 86.49\% Rank-1 accuracy on Market-1501.
With the help of the SP layer, the mAP is increased to 72.51\% and the rank-1 accuracy is increased to 86.52\%.
On DukeMTMC dataset, the mAP and Rank-1 accuracy of the baseline method is 60.83\% and 77.51\% respectively.
While our method increases the mAP to 63.88\%, with 3.05\% gains and Rank-1 accuracy to 79.35\%, with 1.84\% gains, respectively.
Table~\ref{other_datasets} also gives the comparison results on CUHK03 dataset.
The mAP of the baseline method is 49.36\% and the rank-1 accuracy is 53.93\%, our method achieves 51.11\% mAP and 55.86\% rank-1 accuracy.
Note that our method are just based on global features while the state-of-the-art methods on traditional ReID datasets mostly include adequate local feature alignment. Despite of this, our subspace pooling method improves the baselines on all the three datasets.

\begin{table*}[!htbp]
\centering
\fontsize{10}{15}\selectfont
\caption{Results on Market-1501, DukeMTMC and CUHK-03 datasets.}
\label{other_datasets}
       \begin{tabular}{|c|c|c|c|c|c|c|c|c|c|}
    \hline
    \multirow{2}{*}{Dataset}&
    \multicolumn{3}{c|}{Baseline}&
    \multicolumn{3}{c|}{Baseline+SP}&
    \multicolumn{3}{c|}{gains}\cr\cline{2-10}
    &mAP&Rank-1&F-score&mAP&Rank-1&F-score&mAP&Rank-1&F-score\cr
    \hline
    \hline
    Market-1501&71.44&86.49&67.36&72.51&86.52&68.61&\bf{1.07}&\bf{0.03}&\bf{1.25}\cr\hline
    DukeMTMC&60.83&77.51&58.7&63.88&79.35&61.39&\bf{3.05}&\bf{1.84}&\bf{3.39}\cr\hline
    CUHK03&49.36&53.93&49.02&51.11&55.86&51.01&\bf{1.75}&\bf{1.93}&\bf{1.99}\cr\hline
    \end{tabular}
\end{table*}

{\flushleft\textbf{Hyper-parameters.}}
To further investigate how some hyper-parameters affect the final ReID performance, we study the effect of $1\times 1$ convolution layer before the subspace pooling layer.
The original number of feature maps before feeding into $1\times 1$ convolution layer is 2048.
To get a trade-off between efficiency and the accuracy, we set the output of the  $1\times 1$ convolution layer to 512.
The results of different $1\times 1$ convolution layer output sizes are shown in Table~\ref{conv}.

\begin{table}[!htbp]
\centering
\fontsize{10}{15}\selectfont
\caption{The results of different output sizes about 1$\times$1 convolution layer.}
\label{conv}
   \begin{tabular}{c|c|c|c}
    \hline
     size&mAP&Rank-1&F-score\cr
    \hline
    \hline
    64&33.51&43.72&34.39\cr
    \hline
    128&36.41&46.35&36.92\cr
    \hline
    256&37.62&48.24&38.05\cr
    \hline
    512&39.58&49.79&39.82\cr
    \hline
    \end{tabular}
\end{table}

\section{Conclusion}\label{sec_conclusion}
In this paper we collect a large scale airborne person ReID dataset which consists of 39,461 images of 1581 person identities. The images of the dataset are shot by two DJI consumer UAVs flying at
an altitude ranging from 20 to 60 meters above the ground, which covers most of the real UAV surveillance scenarios.
To the best of our knowledge, it is the first large person ReID dataset in aerial imagery facing to the real UAV surveillance scenarios. We hope that our dataset can facilitate the research area of person ReID in aerial imagery or even person ReID on mobile platform.
In addition, we propose to utilize subspace pooling of convolution feature maps to learn a discriminative and compact feature descriptor for ReID in aerial imagery and more importantly, it can be trained via an end-to-end fashion effectively. Extensive experiments show that our method performs favorably against state of the arts on our dataset.

\ifCLASSOPTIONcaptionsoff
  \newpage
\fi



\bibliographystyle{IEEEtran}
\bibliography{IEEEexample}

\begin{thebibliography}{10}
\providecommand{\url}[1]{#1}
\csname url@samestyle\endcsname
\providecommand{\newblock}{\relax}
\providecommand{\bibinfo}[2]{#2}
\providecommand{\BIBentrySTDinterwordspacing}{\spaceskip=0pt\relax}
\providecommand{\BIBentryALTinterwordstretchfactor}{4}
\providecommand{\BIBentryALTinterwordspacing}{\spaceskip=\fontdimen2\font plus
\BIBentryALTinterwordstretchfactor\fontdimen3\font minus
  \fontdimen4\font\relax}
\providecommand{\BIBforeignlanguage}[2]{{%
\expandafter\ifx\csname l@#1\endcsname\relax
\typeout{** WARNING: IEEEtran.bst: No hyphenation pattern has been}%
\typeout{** loaded for the language `#1'. Using the pattern for}%
\typeout{** the default language instead.}%
\else
\language=\csname l@#1\endcsname
\fi
#2}}
\providecommand{\BIBdecl}{\relax}
\BIBdecl

\bibitem{li2014deepreid}
W.~Li, R.~Zhao, T.~Xiao, and X.~Wang, ``Deepreid: Deep filter pairing neural
  network for person re-identification,'' in \emph{Proceedings of the IEEE
  Conference on Computer Vision and Pattern Recognition}, 2014, pp. 152--159.

\bibitem{wang2015zero}
Z.~Wang, R.~Hu, C.~Liang, Y.~Yu, J.~Jiang, M.~Ye, J.~Chen, and Q.~Leng,
  ``Zero-shot person re-identification via cross-view consistency,'' \emph{IEEE
  Transactions on Multimedia}, vol.~18, no.~2, pp. 260--272, 2015.

\bibitem{ye2016person}
M.~Ye, C.~Liang, Y.~Yu, Z.~Wang, Q.~Leng, C.~Xiao, J.~Chen, and R.~Hu, ``Person
  reidentification via ranking aggregation of similarity pulling and
  dissimilarity pushing,'' \emph{IEEE Transactions on Multimedia}, vol.~18,
  no.~12, pp. 2553--2566, 2016.

\bibitem{xiao2016learning}
T.~Xiao, H.~Li, W.~Ouyang, and X.~Wang, ``Learning deep feature representations
  with domain guided dropout for person re-identification,'' in
  \emph{Proceedings of the IEEE Conference on Computer Vision and Pattern
  Recognition}, 2016, pp. 1249--1258.

\bibitem{zheng2016person}
L.~Zheng, Y.~Yang, and A.~G. Hauptmann, ``Person re-identification: Past,
  present and future,'' \emph{arXiv preprint arXiv:1610.02984}, 2016.

\bibitem{hermans2017defense}
A.~Hermans, L.~Beyer, and B.~Leibe, ``In defense of the triplet loss for person
  re-identification,'' \emph{arXiv preprint arXiv:1703.07737}, 2017.

\bibitem{zhou2017large}
S.~Zhou, J.~Wang, R.~Shi, Q.~Hou, Y.~Gong, and N.~Zheng, ``Large margin
  learning in set-to-set similarity comparison for person reidentification,''
  \emph{IEEE Transactions on Multimedia}, vol.~20, no.~3, pp. 593--604, 2017.

\bibitem{zhang2018pedestrian}
S.~Zhang, D.~Cheng, Y.~Gong, D.~Shi, X.~Qiu, Y.~Xia, and Y.~Zhang, ``Pedestrian
  search in surveillance videos by learning discriminative deep features,''
  \emph{Neurocomputing}, vol. 283, pp. 120--128, 2018.

\bibitem{chen2018person}
Y.~Chen, X.~Zhu, S.~Gong \emph{et~al.}, ``Person re-identification by deep
  learning multi-scale representations,'' 2018.

\bibitem{wang2019incremental}
Z.~Wang, J.~Jiang, Y.~Yu, and S.~Satoh, ``Incremental re-identification by
  cross-direction and cross-ranking adaption,'' \emph{IEEE Transactions on
  Multimedia}, 2019.

\bibitem{ding2018feature}
G.~Ding, S.~Zhang, S.~Khan, Z.~Tang, J.~Zhang, and F.~Porikli, ``Feature
  affinity based pseudo labeling for semi-supervised person
  re-identification,'' \emph{arXiv preprint arXiv:1805.06118}, 2018.

\bibitem{zhang2019person}
S.~Zhang, R.~Cao, X.~Wei, P.~Wang, and Y.~Zhang, ``Person re-identification
  with neural architecture search,'' in \emph{Chinese Conference on Pattern
  Recognition and Computer Vision (PRCV)}.\hskip 1em plus 0.5em minus
  0.4em\relax Springer, 2019, pp. 540--551.

\bibitem{diangang2020}
D.~Li, X.~Wei, X.~Hong, and Y.~Gong, ``Infrared-visible cross-modal person
  re-identification with an x modality,'' in \emph{The Thirty-Fourth {AAAI}
  Conference on Artificial Intelligence}, 2020.

\bibitem{xia2018dota}
G.-S. Xia, X.~Bai, J.~Ding, Z.~Zhu, S.~Belongie, J.~Luo, M.~Datcu, M.~Pelillo,
  and L.~Zhang, ``Dota: A large-scale dataset for object detection in aerial
  images,'' in \emph{Proc. CVPR}, 2018.

\bibitem{zhu2018vision}
P.~Zhu, L.~Wen, X.~Bian, H.~Ling, and Q.~Hu, ``Vision meets drones: a
  challenge,'' \emph{arXiv preprint arXiv:1804.07437}, 2018.

\bibitem{mueller2016benchmark}
M.~Mueller, N.~Smith, and B.~Ghanem, ``A benchmark and simulator for uav
  tracking,'' in \emph{European conference on computer vision}.\hskip 1em plus
  0.5em minus 0.4em\relax Springer, 2016, pp. 445--461.

\bibitem{cheng2016learning}
G.~Cheng, P.~Zhou, and J.~Han, ``Learning rotation-invariant convolutional
  neural networks for object detection in vhr optical remote sensing images,''
  \emph{IEEE Transactions on Geoscience and Remote Sensing}, vol.~54, no.~12,
  pp. 7405--7415, 2016.

\bibitem{zhou2018scale}
P.~Zhou, B.~Ni, C.~Geng, J.~Hu, and Y.~Xu, ``Scale-transferrable object
  detection,'' in \emph{Proceedings of the IEEE Conference on Computer Vision
  and Pattern Recognition}, 2018, pp. 528--537.

\bibitem{zhang2018w2f}
Y.~Zhang, Y.~Bai, M.~Ding, Y.~Li, and B.~Ghanem, ``W2f: A weakly-supervised to
  fully-supervised framework for object detection,'' in \emph{Proceedings of
  the IEEE Conference on Computer Vision and Pattern Recognition}, 2018, pp.
  928--936.

\bibitem{zhou2018learning}
P.~Zhou, X.~Han, V.~I. Morariu, and L.~S. Davis, ``Learning rich features for
  image manipulation detection,'' in \emph{Proceedings of the IEEE Conference
  on Computer Vision and Pattern Recognition}, 2018, pp. 1053--1061.

\bibitem{xiang2014monocular}
Y.~Xiang, C.~Song, R.~Mottaghi, and S.~Savarese, ``Monocular multiview object
  tracking with 3d aspect parts,'' in \emph{European Conference on Computer
  Vision}.\hskip 1em plus 0.5em minus 0.4em\relax Springer, 2014, pp. 220--235.

\bibitem{sun2017svdnet}
Y.~Sun, L.~Zheng, W.~Deng, and S.~Wang, ``Svdnet for pedestrian retrieval,''
  \emph{arXiv preprint}, vol.~1, no.~6, 2017.

\bibitem{Wei_2018_CVPR}
X.~Wei, Y.~Zhang, Y.~Gong, and N.~Zheng, ``Kernelized subspace pooling for deep
  local descriptors,'' in \emph{The IEEE Conference on Computer Vision and
  Pattern Recognition (CVPR)}, June 2018.

\bibitem{wei2018grassmann}
X.~Wei, Y.~Zhang, Y.~Gong, J.~Zhang, and N.~Zheng, ``Grassmann pooling as
  compact homogeneous bilinear pooling for fine-grained visual
  classification,'' in \emph{Proceedings of the European Conference on Computer
  Vision (ECCV)}, 2018, pp. 355--370.

\bibitem{jayasumana2015kernel}
S.~Jayasumana, R.~Hartley, M.~Salzmann, H.~Li, and M.~Harandi, ``Kernel methods
  on riemannian manifolds with gaussian rbf kernels,'' \emph{IEEE transactions
  on pattern analysis and machine intelligence}, vol.~37, no.~12, pp.
  2464--2477, 2015.

\bibitem{singh2018eye}
A.~Singh, D.~Patil, and S.~Omkar, ``Eye in the sky: Real-time drone
  surveillance system (dss) for violent individuals identification using
  scatternet hybrid deep learning network,'' in \emph{Proceedings of the IEEE
  Conference on Computer Vision and Pattern Recognition Workshops}, 2018, pp.
  1629--1637.

\bibitem{zheng2016mars}
L.~Zheng, Z.~Bie, Y.~Sun, J.~Wang, C.~Su, S.~Wang, and Q.~Tian, ``Mars: A video
  benchmark for large-scale person re-identification,'' in \emph{European
  Conference on Computer Vision}.\hskip 1em plus 0.5em minus 0.4em\relax
  Springer, 2016, pp. 868--884.

\bibitem{zheng2017unlabeled}
Z.~Zheng, L.~Zheng, and Y.~Yang, ``Unlabeled samples generated by gan improve
  the person re-identification baseline in vitro,'' in \emph{Proceedings of the
  IEEE International Conference on Computer Vision}, 2017, pp. 3754--3762.

\bibitem{layne2014investigating}
R.~Layne, T.~M. Hospedales, and S.~Gong, ``Investigating open-world person
  re-identification using a drone,'' in \emph{European Conference on Computer
  Vision}.\hskip 1em plus 0.5em minus 0.4em\relax Springer, 2014, pp. 225--240.

\bibitem{Zheng2016Zero}
W.~Zheng, R.~Hu, L.~Chao, Y.~Yi, and Q.~Leng, ``Zero-shot person
  re-identification via cross-view consistency,'' \emph{IEEE Transactions on
  Multimedia}, vol.~18, no.~2, pp. 260--272, 2016.

\bibitem{Alam2017A}
M.~Alam, M.~Bennamoun, R.~Togneri, and F.~Sohel, ``A joint deep boltzmann
  machine (jdbm) model for person identification using mobile phone data,''
  \emph{IEEE Transactions on Multimedia}, vol.~PP, no.~99, pp. 1--1, 2017.

\bibitem{ahmed2015improved}
E.~Ahmed, M.~Jones, and T.~K. Marks, ``An improved deep learning architecture
  for person re-identification,'' in \emph{Proceedings of the IEEE Conference
  on Computer Vision and Pattern Recognition}, 2015, pp. 3908--3916.

\bibitem{liu2015set}
H.~Liu, B.~Ma, L.~Qin, J.~Pang, C.~Zhang, and Q.~Huang, ``Set-label modeling
  and deep metric learning on person re-identification,''
  \emph{Neurocomputing}, vol. 151, pp. 1283--1292, 2015.

\bibitem{yi2014deep}
D.~Yi, Z.~Lei, S.~Liao, and S.~Z. Li, ``Deep metric learning for person
  re-identification,'' in \emph{Pattern Recognition (ICPR), 2014 22nd
  International Conference on}.\hskip 1em plus 0.5em minus 0.4em\relax IEEE,
  2014, pp. 34--39.

\bibitem{liu2018part}
C.~Liu, T.~Bao, and M.~Zhu, ``Part-based feature extraction for person
  re-identification,'' in \emph{Proceedings of the 2018 10th International
  Conference on Machine Learning and Computing}.\hskip 1em plus 0.5em minus
  0.4em\relax ACM, 2018, pp. 172--177.

\bibitem{xiang2018homocentric}
W.~Xiang, J.~Huang, X.~Qi, X.~Hua, and L.~Zhang, ``Homocentric hypersphere
  feature embedding for person re-identification,'' \emph{arXiv preprint
  arXiv:1804.08866}, 2018.

\bibitem{Ge2018FD}
Y.~Ge, Z.~Li, H.~Zhao, G.~Yin, S.~Yi, X.~Wang, and H.~Li, ``Fd-gan: Pose-guided
  feature distilling gan for robust person re-identification,'' 2018.

\bibitem{li2018unsupervised}
M.~Li, X.~Zhu, and S.~Gong, ``Unsupervised person re-identification by deep
  learning tracklet association,'' \emph{arXiv preprint arXiv:1809.02874},
  2018.

\bibitem{Yu2019Unsupervised}
H.~X. Yu, W.~S. Zheng, A.~Wu, X.~Guo, and J.~H. Lai, ``Unsupervised person
  re-identification by soft multilabel learning,'' 2019.

\bibitem{shen2018person}
Y.~Shen, H.~Li, S.~Yi, D.~Chen, and X.~Wang, ``Person re-identification with
  deep similarity-guided graph neural network,'' in \emph{European Conference
  on Computer Vision}.\hskip 1em plus 0.5em minus 0.4em\relax Springer, 2018,
  pp. 508--526.

\bibitem{wei2018person}
L.~Wei, S.~Zhang, W.~Gao, and Q.~Tian, ``Person transfer gan to bridge domain
  gap for person re-identification,'' in \emph{Proceedings of the IEEE
  Conference on Computer Vision and Pattern Recognition}, 2018, pp. 79--88.

\bibitem{li2018harmonious}
W.~Li, X.~Zhu, and S.~Gong, ``Harmonious attention network for person
  re-identification,'' in \emph{CVPR}, vol.~1, 2018, p.~2.

\bibitem{sun2018dissecting}
X.~Sun and L.~Zheng, ``Dissecting person re-identification from the viewpoint
  of viewpoint,'' \emph{arXiv preprint arXiv:1812.02162}, 2018.

\bibitem{xie2017all}
D.~Xie, J.~Xiong, and S.~Pu, ``All you need is beyond a good init: Exploring
  better solution for training extremely deep convolutional neural networks
  with orthonormality and modulation,'' \emph{arXiv preprint arXiv:1703.01827},
  2017.

\bibitem{chan2015pcanet}
T.-H. Chan, K.~Jia, S.~Gao, J.~Lu, Z.~Zeng, and Y.~Ma, ``Pcanet: A simple deep
  learning baseline for image classification?'' \emph{IEEE transactions on
  image processing}, vol.~24, no.~12, pp. 5017--5032, 2015.

\bibitem{sun2018beyond}
Y.~Sun, L.~Zheng, Y.~Yang, Q.~Tian, and S.~Wang, ``Beyond part models: Person
  retrieval with refined part pooling (and a strong convolutional baseline),''
  in \emph{Proceedings of the European Conference on Computer Vision (ECCV)},
  2018, pp. 480--496.

\bibitem{zhang2017alignedreid}
X.~Zhang, H.~Luo, X.~Fan, W.~Xiang, Y.~Sun, Q.~Xiao, W.~Jiang, C.~Zhang, and
  J.~Sun, ``Alignedreid: Surpassing human-level performance in person
  re-identification,'' \emph{arXiv preprint arXiv:1711.08184}, 2017.

\bibitem{ustinova2017multi}
E.~Ustinova, Y.~Ganin, and V.~Lempitsky, ``Multi-region bilinear convolutional
  neural networks for person re-identification,'' in \emph{2017 14th IEEE
  International Conference on Advanced Video and Signal Based Surveillance
  (AVSS)}.\hskip 1em plus 0.5em minus 0.4em\relax IEEE, 2017, pp. 1--6.

\bibitem{zheng2018discriminatively}
Z.~Zheng, L.~Zheng, and Y.~Yang, ``A discriminatively learned cnn embedding for
  person reidentification,'' \emph{ACM Transactions on Multimedia Computing,
  Communications, and Applications (TOMM)}, vol.~14, no.~1, p.~13, 2018.

\bibitem{zhao2017deeply}
L.~Zhao, X.~Li, Y.~Zhuang, and J.~Wang, ``Deeply-learned part-aligned
  representations for person re-identification,'' in \emph{Proceedings of the
  IEEE International Conference on Computer Vision}, 2017, pp. 3219--3228.

\bibitem{he2018deep}
L.~He, J.~Liang, H.~Li, and Z.~Sun, ``Deep spatial feature reconstruction for
  partial person re-identification: Alignment-free approach,'' in
  \emph{Proceedings of the IEEE Conference on Computer Vision and Pattern
  Recognition}, 2018, pp. 7073--7082.

\bibitem{zhong2018camera}
Z.~Zhong, L.~Zheng, Z.~Zheng, S.~Li, and Y.~Yang, ``Camera style adaptation for
  person re-identification,'' in \emph{Proceedings of the IEEE Conference on
  Computer Vision and Pattern Recognition}, 2018, pp. 5157--5166.

\bibitem{wang2018learning}
G.~Wang, Y.~Yuan, X.~Chen, J.~Li, and X.~Zhou, ``Learning discriminative
  features with multiple granularities for person re-identification,'' in
  \emph{2018 ACM Multimedia Conference on Multimedia Conference}.\hskip 1em
  plus 0.5em minus 0.4em\relax ACM, 2018, pp. 274--282.

\bibitem{zhou2019omni}
K.~Zhou, Y.~Yang, A.~Cavallaro, and T.~Xiang, ``Omni-scale feature learning for
  person re-identification,'' \emph{arXiv preprint arXiv:1905.00953}, 2019.

\end{thebibliography}

\end{document}